%% file: acl_latex.tex
\documentclass[11pt]{article}

\usepackage[preprint]{acl}
\usepackage{amsmath}
\usepackage{times}
\usepackage{amsfonts}
\usepackage{latexsym}
\usepackage{booktabs}
\usepackage[T1]{fontenc}
\usepackage{etoc}

\usepackage[utf8]{inputenc}
\usepackage{xcolor}
\usepackage{epigraph}
\usepackage{microtype}
\usepackage{caption}
\usepackage{tabularx}
\usepackage{booktabs}
\usepackage[table]{xcolor}
\usepackage{amsmath}
\usepackage{inconsolata}

\usepackage{graphicx}
\usepackage{soul}
\usepackage{cuted}
\usepackage{amssymb}
\usepackage{hyperref}

\usepackage[most]{tcolorbox}
\usepackage{xcolor}
\usepackage{array} 
\definecolor{quanback}{RGB}{248,249,252} 
\definecolor{quanframe}{RGB}{107,136,171} 
\definecolor{judgeback}{RGB}{250,247,243}
\definecolor{judgeframe}{RGB}{198,120,66}

\newtcblisting{promptbox}[2][]{%
  breakable,
  enhanced,
  listing only,
  colback=quanback,
  colframe=quanframe,
  coltitle=black,
  fonttitle=\bfseries\large,
  title={#2},
  sharp corners,
  boxrule=0.7pt,
  left=1.2mm,right=1.2mm,top=1.0mm,bottom=1.0mm,
  listing options={basicstyle=\ttfamily\small,breaklines=true},
  #1
}

\newtcblisting{judgebox}[2][]{%
  breakable,
  enhanced,
  listing only,
  colback=judgeback,
  colframe=judgeframe,
  coltitle=black,
  fonttitle=\bfseries\large,
  title={#2},
  sharp corners,
  boxrule=0.7pt,
  left=1.2mm,right=1.2mm,top=1.0mm,bottom=1.0mm,
  listing options={basicstyle=\ttfamily\small,breaklines=true},
  #1
}

\newcommand{\apptocheading}[1]{%
  \par\vspace{3.5ex plus 1ex minus .2ex}%
  {\normalfont\Large\bfseries #1\par}%
  \vspace{2.3ex plus .2ex}%
}

\newcommand{\apptocline}[3]{%
  \noindent\hspace*{#1}%
  \begin{tabularx}{\dimexpr\linewidth-#1\relax}{@{}>{\raggedright\arraybackslash}X@{\dotfill}r@{}}%
    \hyperref[#2]{#3} & \hyperref[#2]{\pageref*{#2}}%
  \end{tabularx}\par
}

\newcommand{\apptocsection}[2]{%
  \vspace{0.25em}%
  \apptocline{0em}{#1}{\textbf{#2}}%
}

\newcommand{\apptocsubsection}[2]{%
  \apptocline{1.5em}{#1}{#2}%
}

\title{\alias: Benchmarking \underline{Sci}entific \underline{I}nstruction \underline{F}ollowing \\ Towards Rigorous Scientific Intelligence}

\author{
  \parbox{\textwidth}{\centering
  Encheng Su\textsuperscript{1,2,\dag},
  Jianyu Wu\textsuperscript{1,3,\dag},
  Lintao Wang\textsuperscript{1,5},
  Pengze Li\textsuperscript{1,6},\\
  Aoran Wang\textsuperscript{1},
  Jinouwen Zhang\textsuperscript{1},
  Yizhou Wang\textsuperscript{1,4},
  Yuan Meng\textsuperscript{7},\\
  Chen Tang\textsuperscript{1,4,*},
  Xinzhu Ma\textsuperscript{1,8,*},
  Shixiang Tang\textsuperscript{1,4,*},
  Houqiang Li\textsuperscript{2,*}
  \\[0.35em]
  {\small
  \textsuperscript{1}Shanghai AI Laboratory \quad
  \textsuperscript{2}University of Science and Technology of China \quad
  \textsuperscript{3}Shanghai Jiao Tong University \quad \\
  \textsuperscript{4}The Chinese University of Hong Kong \quad
  \textsuperscript{5}University of Sydney \quad
  \textsuperscript{6}Fudan University\quad \\
  \textsuperscript{7}Tsinghua University \quad
  \textsuperscript{8}Beihang University
  }
  \\[0.35em]
  {\small\textsuperscript{\dag}Equal contribution. \quad \textsuperscript{*}Corresponding authors.}
  }
}

\newcommand{\alias}{{SciIF}}

\begin{document}
\maketitle

% \begin{figure*}[t]
%     \centering
%     \includegraphics[width=1\linewidth]{figures/intro.pdf}
%     \caption{(a) Limitations of answer-only scientific benchmarks: they score final-answer correctness but do not check scientific constraints. (b) \alias (Sci-IF) expands evaluation with a constraint catalog and auditable, evidence-based verification; we show benchmark statistics and a boundary-condition example verified by substitution checks. (c) Training on Sci-IF: SFT improves correctness, while RL further boosts explicit, auditable constraint compliance; we illustrate an applicability-range case and schematic gains on IFEval and MMLU (science).
% }
%     \label{fig:intro_example}
% \end{figure*}
\vspace{-0.8em}
\begin{strip}
\centering
\includegraphics[width=\textwidth]{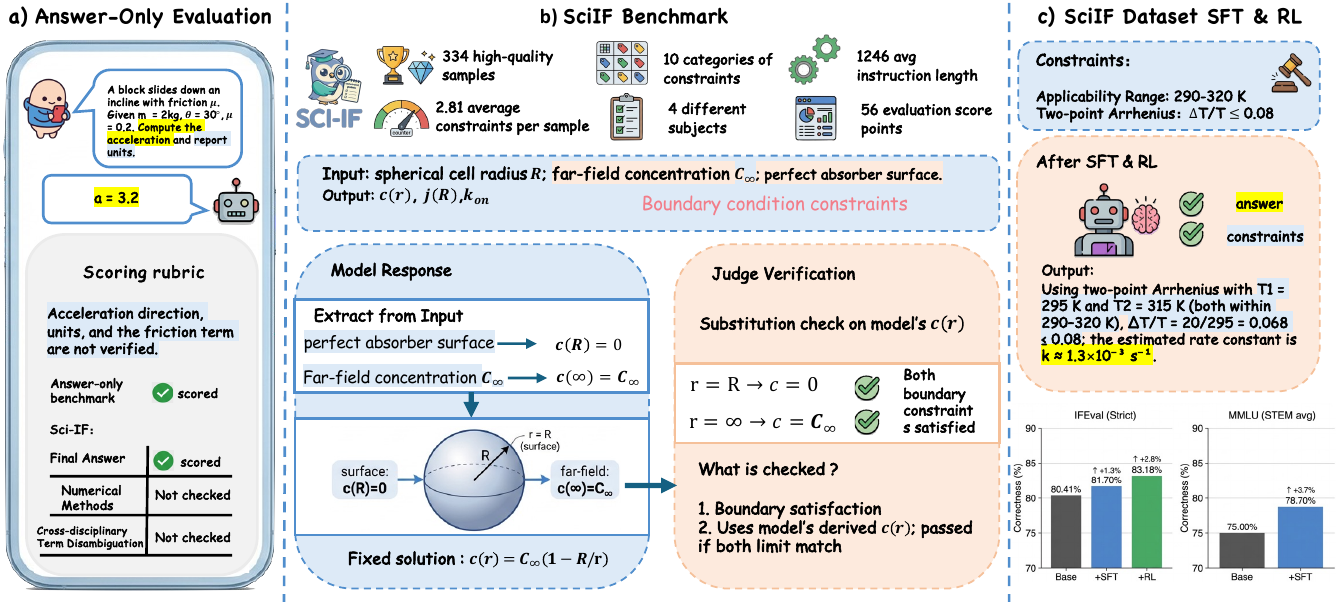}
\captionof{figure}{%
Overview of the \alias{} Benchmark. 
a) Existing answer-only scientific benchmarks evaluate only final answer matching, overlooking scientific constraints and instruction compliance.
b) \alias{} advances this paradigm by introducing explicit scientific constraints and structured verification rubrics, enabling fine-grained assessment of scientific constraint-aware reasoning.
c) \alias{} provides constraint-grounded SFT and RL datasets, showing that scientific instruction following capability systematically improves correctness, rigor, and instruction adherence.
}
\label{fig:teaser}
\end{strip}
\vspace{-1.0em}

\begin{abstract}
As large language models (LLMs) transition from general knowledge retrieval to complex scientific discovery, their evaluation standards must also incorporate the rigorous norms of scientific inquiry. 
Existing benchmarks exhibit a critical blind spot: general instruction-following metrics focus on superficial formatting, while domain-specific scientific benchmarks assess only final-answer correctness, often rewarding models that arrive at the right result with the wrong reasons. 
To address this gap, we introduce scientific instruction following: the capability to solve problems while strictly adhering to the constraints that establish scientific validity. 
Specifically, we introduce \alias, a multi-discipline benchmark that evaluates this capability by pairing university-level problems with a fixed catalog of constraints across three pillars: scientific conditions (\emph{e.g.}, boundary checks and assumptions), semantic stability (\emph{e.g.}, unit and symbol conventions), and specific processes (\emph{e.g.}, required numerical methods). Uniquely, \alias~emphasizes auditability, requiring models to provide explicit evidence of constraint satisfaction rather than implicit compliance. 
By measuring both solution correctness and multi-constraint adherence, \alias~enables fine-grained diagnosis of compositional reasoning failures, ensuring that LLMs can function as reliable agents within the strict logical frameworks of science. Data and code are available at https://github.com/suencgo/SciIF.git.
\end{abstract}

\input{section/intro}
\input{section/related-work}

\input{section/method}

\input{section/experiments}

\section{Conclusion}
We introduce \alias~to evaluate scientific instruction following, addressing the limitations of current benchmarks that overlook the interplay between solution correctness and constraint satisfaction. Through a taxonomy of domain-specific constraints and an auditable evaluation protocol, \alias~distinguishes between rigorous reasoning and lucky guessing. 
In addition, our protocol emphasizes transparency and reproducibility, enabling consistent comparisons across models and settings.
Crucially, we show that the rigor demanded by \alias~generalizes: models fine-tuned on our data exhibit dual improvements, boosting IFEval by 2.8 points and MMLU-Physics by 8.0 points. We will release \alias~to facilitate the development of LLMs that are not only knowledgeable but strictly compliant with the norms of scientific practice.

\section*{Limitation}

We acknowledge that our dataset focuses on text-based problems, omitting multimodal scientific tasks (e.g., DNA, RNA sequence generation/understanding). Future work will extend \alias~to multimodal settings and explore iterative, multi-turn scientific agents.

% \section*{Acknowledgments}

% This document has been adapted
% by Steven Bethard, Ryan Cotterell and Rui Yan
% from the instructions for earlier ACL and NAACL proceedings, including those for
% ACL 2019 by Douwe Kiela and Ivan Vuli\'{c},
% NAACL 2019 by Stephanie Lukin and Alla Roskovskaya,
% ACL 2018 by Shay Cohen, Kevin Gimpel, and Wei Lu,
% NAACL 2018 by Margaret Mitchell and Stephanie Lukin,
% Bib\TeX{} suggestions for (NA)ACL 2017/2018 from Jason Eisner,
% ACL 2017 by Dan Gildea and Min-Yen Kan,
% NAACL 2017 by Margaret Mitchell,
% ACL 2012 by Maggie Li and Michael White,
% ACL 2010 by Jing-Shin Chang and Philipp Koehn,
% ACL 2008 by Johanna D. Moore, Simone Teufel, James Allan, and Sadaoki Furui,
% ACL 2005 by Hwee Tou Ng and Kemal Oflazer,
% ACL 2002 by Eugene Charniak and Dekang Lin,
% and earlier ACL and EACL formats written by several people, including
% John Chen, Henry S. Thompson and Donald Walker.
% Additional elements were taken from the formatting instructions of the \emph{International Joint Conference on Artificial Intelligence} and the \emph{Conference on Computer Vision and Pattern Recognition}.

\bibliography{custom}

\clearpage
\newpage
\appendix
\input{section/appendix}

\end{document}

%% file: section/intro.tex
\section{Introduction}
\label{sec:intro}

\epigraph{\textit{``What we observe is not nature itself, but nature exposed to our method of questioning.''}}{--- Werner Heisenberg}

With the rapid adoption of large language models (LLMs) in scientific discovery, the expectation for model capability is shifting from simple knowledge retrieval to complex problem solving~\cite{hu2025survey}. However, a fundamental gap exists in how we evaluate these foundation models: in scientific context, the validity of an answer is not intrinsic to the result itself, but is contingent upon whether it is produced within a specific framework of constraints. Unlike everyday user requests where a helpful answer is sufficient, scientific problems are governed by strict norms within the inputs (\emph{e.g.}, assumptions, boundary conditions, definitions, and procedures) that dictate whether a solution is scientifically meaningful or merely numerically plausible. Therefore, as models increasingly tackle rigor-grounded scientific tasks~\citep{mitchener2025kosmos,wu2025bigbang,wang2025scireasoner}, evaluation must go beyond answer plausibility to assess whether they can strictly adhere to the constraints that establish the scientific validity of that result, namely scientific instruction-following. 

However, current benchmarks fail to capture the critical interplay between scientific solutions and justifying constraints. 
Mainstream instruction-following benchmarks \citep{zhou2023ifeval,jiang2024followbench,zhang2025cfbench,qin2024infobench} primarily emphasize surface-level \emph{format compliance}. For example, IFEval~\cite{zhou2023ifeval} evaluates properties like word counts or JSON validity. While effective for general-purpose dialogue, these criteria are insufficient for science tasks. A model may perfectly satisfy formatting rules yet violating critical scientific rigor, such as applying a formula where it is mathematically undefined or conflating incompatible unit systems. 
Conversely, prevailing scientific benchmarks \citep{hendrycks2021mmlu,wang2023scibench} focus almost exclusively on \emph{final-answer correctness}. By treating the reasoning and constraint sanctification process as a black box, these evaluation systematically overlook ``right-for-the-wrong-reasons'' failures, failing to distinguish a rigorous scientific agent from one that merely arrives at the correct answer without adhering to the underlying scientific framework.

We address this issue by first defining \emph{scientific instruction following}: the ability to solve a scientific problem while explicitly satisfying the constraints that govern correctness, meaning, and process. Unlike general instruction following, which often concerns stylistic preferences and fluency~\citep{wei2022emergent}, this capability rests on three pillars of scientific validity. 
First, it requires adherence to \emph{scientific conditions}: models must explicitly check applicability ranges, verify boundary conditions, and state assumptions to prove the solution is valid under the given parameters~\citep{ribeiro2020beyond}. Second, it demands \emph{semantic stability}: models must rigorously adhere to unit conventions, symbol definitions, and terminology standards to prevent “meaning drift”, where a model implicitly substitutes one quantity for another. Third, it necessitates adherence to specific \emph{scientific processes}: when a specific numerical method or experimental protocol is requested, the model must provide auditable, executable steps rather than a generic conclusion.

In this paper, we introduce \alias, a benchmark for scientific instruction following across multiple disciplines. Each instance contains: (i) a university-level scientific problem, (ii) an enabled subset of constraints drawn from a predefined catalog, and (iii) a reference solution that is validated to be correct and consistent with the enabled constraints. The catalog is designed around scientific practice and spans three families. \emph{Condition constraints} regulate modeling validity (\emph{e.g.}, assumptions, boundary conditions, applicability range, unit conventions). \emph{Terminology constraints} prevent meaning drift (\emph{e.g.}, cross-disciplinary disambiguation, in-domain term definitions, symbol and constant conventions, variable naming consistency). \emph{Process constraints} require a specific scientific procedure (\emph{e.g.}, numerical methods or experimental methods). Because constraints are drawn from a fixed catalog and combined in controlled ways, \alias~ can test compositional instruction following: whether models can coordinate multiple scientific requirements in a single coherent answer \citep{lake2018generalization}.

A central design goal of \alias~ is \emph{auditability}. For each enabled constraint, we specify required evidence that must appear in the model output; judges are instructed not to infer missing evidence. This makes compliance decisions depend on what the model explicitly commits to, rather than on judge guesswork. We report answer correctness, overall multi-constraint compliance, and per-constraint pass rates, enabling fine-grained diagnosis of where scientific instruction following breaks. 
Furthermore, we demonstrate that fine-tuning on \alias~ confers dual benefits, delivering a 2.8\% boost on general instruction following (IFEval) and a remarkable 8.0\% improvement on scientific tasks (MMLU-Physics).
This indicates that the rigor required for scientific constraints generalizes effectively, sharpening the model's ability to follow complex rules while deepening its domain expertise.

In summary, our contributions are as follows:
\begin{itemize}
\item We introduce \alias, a multi-discipline benchmark that shifts the evaluation focus from mere numerical correctness to the rigorous execution of scientific instructions.
\item We construct a predefined catalog of constraints that supports the generation of controlled problem mixtures for fine-grained failure diagnosis.
\item We propose an auditable, evidence-based evaluation protocol that ensures compliance is measured by explicit logical commitments rather than implicit guesswork.
\item We demonstrate the utility of \alias~ for model alignment, showing that learning strict scientific constraints generalizes to improve both general instruction following and domain-specific reasoning.
\end{itemize}

% In summary, \alias~ makes the following contributions: 
% \begin{itemize}
%     \item A multi-discipline benchmark that evaluates scientific instruction following beyond final-answer correctness.
%     \item A fixed catalog of scientific constraints that supports controlled mixtures and diagnostic analysis.
%     \item An evidence-based evaluation protocol that yields auditable compliance scores and interpretable failure modes.
%     \item Empirical evidence demonstrating that fine-tuning on \alias~ confers transferable gains, improving both general instruction-following capabilities and performance on established scientific QA tasks. 
    
% \end{itemize}
%  % (2) ; (3) s; and (4) 

%% file: section/related-work.tex
\section{Related Work}
\label{sec:related}

% \paragraph{Answer-centric benchmarks and the missing compliance axis.}
\paragraph{Answer-Centric Scientific Benchmarks.}
Most scientific and academic evaluations score models primarily by \emph{final-answer correctness}. This includes both \emph{single-discipline} benchmarks such as PhysUniBench \cite{wang2025physunibenchundergraduatelevelphysicsreasoning} that target a specific domain and \emph{multi-discipline} suites---such as MMLU \cite{hendrycks2021mmlu} and BIG-Bench Hard \cite{suzgun2022bbh}---as well as science-focused evaluations such as SciBench \cite{wang2023scibench} and SciEvalKit\cite{wang2026scievalkitopensourceevaluationtoolkit}. While these benchmarks effectively quantify knowledge and reasoning under objective metrics, they often treat the derivation process as a black box. As a result, models can be ``right for the wrong reasons'': producing a correct value while violating constraints that govern scientific validity and meaning (e.g., unstated assumptions, missing applicability checks, unit inconsistencies, or symbol drift). \alias~complements correctness-centric evaluation by explicitly measuring scientific constraint compliance as a separate axis. Concretely, it scores whether required scientific evidence is \emph{stated and checked} in the output (e.g., explicit boundary substitutions or applicability-range validation), rather than assuming that a plausible derivation implicitly satisfies the rules.

% \paragraph{From surface rules to scientific constraint auditing.}
% A related direction evaluates instruction following via automatically verifiable constraints. IFEval focuses on surface-form requirements such as word counts, required keywords, and formatting templates \cite{zhou2023ifeval}. In contrast, \alias targets \emph{semantically grounded} scientific constraints that must be supported by explicit, problem-grounded evidence (e.g., boundary-condition verification, applicability-range statements, unit discipline, and stable symbol/terminology usage), aiming to make compliance decisions auditable rather than inferred. This evidence-first design supports fine-grained diagnosis: we can attribute failures to specific constraint families (conditions, semantic stability, or process), instead of collapsing them into a single correctness score. 

\paragraph{General Instruction Following Evaluation.}
Instruction following has been studied as cross-task generalization from natural language instructions \cite{mishra-etal-2022-cross,sanh2022multitask,zhou2023ifeval,jiang2024followbench,zhang2025cfbench,qin2024infobench}, and improved through instruction tuning and synthetic instruction generation \cite{wei2022flan,chung2022scaling,wang-etal-2023-self-instruct}. Human-centric evaluation and LLM-as-a-judge frameworks optimize for preference and overall assistant quality \cite{ouyang2022training,bai2022hh,bai2022constitutional,dubois2023alpacafarm,alpacaeval2023,zheng2023judging,chiang2023vicuna}, but they typically do not require explicit evidence that scientific constraints are satisfied; fluent outputs can appear acceptable even when scientific conventions are violated. \alias~is complementary: it evaluates whether models both solve the task and explicitly satisfy the scientific constraints that justify the solution, and it enables tracking trade-offs between correctness and constraint compliance under controlled mixtures of constraints.

%% file: section/method.tex
\section{Benchmark Construction}
\label{sec:benchmark_construction}

\begin{figure*}
    \centering
    \includegraphics[width=1\linewidth]{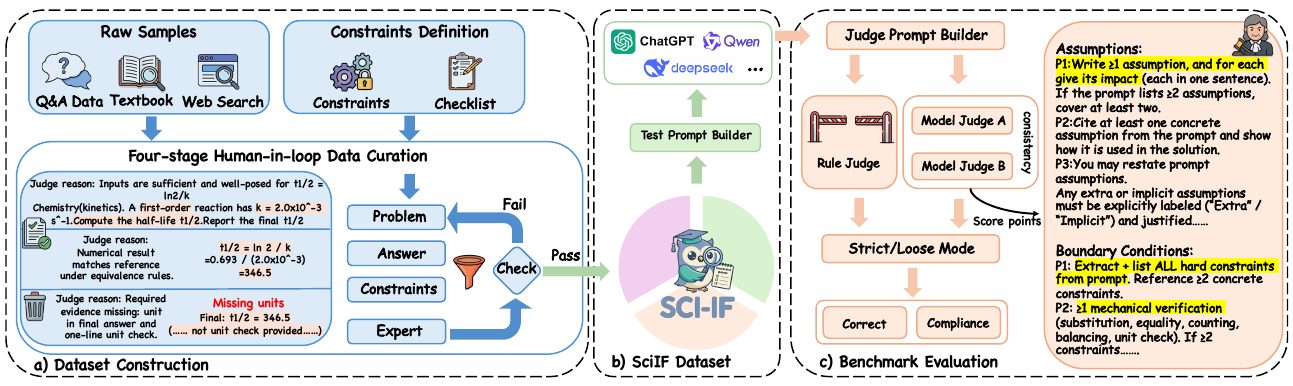}
    \caption{Overview of \alias. We curate scientific QA data from multiple sources and apply a four-stage human-in-the-loop process to produce well-posed problems paired with explicit scientific constraints and auditable evidence checklists. For evaluation, prompt builders generate answer-generation and per-constraint judge prompts, and two independent model judges audit both answer correctness and constraint compliance under Strict or Loose policies. The example highlights a core failure mode: an answer can match the reference numerically yet fail compliance when required evidence such as units and a one-line unit check is missing.}
    \label{fig:placeholder}
\end{figure*}

\alias\ is designed to evaluate \emph{scientific instruction following} as a capability distinct from answer accuracy. In scientific problem solving, constraints such as unit discipline, symbol meaning, validity conditions, and method requirements are not cosmetic---they determine whether an answer is interpretable, reproducible, and even semantically correct. To surface these failures, \alias\ separates evaluation into two axes: (i) correctness of the scientific outcome, and (ii) explicit, auditable compliance with enabled scientific constraints.

\subsection{Task Definition}
\label{sec:task_def}

\alias\ evaluates scientific instruction following along two separate axes.
Given a scientific problem, we score not only whether a model reaches the correct scientific result, but also whether it explicitly follows the enabled scientific constraints.

Each instance is a triple $(x, C, y^\star)$.
Here $x$ is a university-level scientific problem, $C$ is a small set of enabled constraints drawn from a fixed catalog, and $y^\star$ is a reference solution that is both correct and compliant with $C$.
A model receives $(x, C)$ and produces an output $\hat{y}$.

\paragraph{Answer Correctness.}
We score whether $\hat{y}$ matches $y^\star$ on the required scientific outcome, using task-appropriate equivalence criteria.

\paragraph{Constraint Compliance.}
In parallel, we score whether $\hat{y}$ provides explicit, problem-grounded evidence for each enabled constraint in $C$.
Judges check evidence presence rather than infer missing assumptions or validity checks, which cleanly separates \emph{correct-but-non-compliant} outputs from \emph{compliant-but-incorrect} ones.

\subsection{Dataset and Constraint Catalog}
\label{sec:benchmark_design}

\alias\ uses a fixed test set of 334 university-level problems across Biology, Chemistry, Materials, and Physics, and we additionally release 910 training problems used for SFT and verifier-based RL.
Each test instance enables a small set of constraints drawn from a fixed catalog of ten atomic constraints.
The catalog spans three families: condition constraints (Assumptions, Boundary Conditions, Applicability Range, Units Standard), terminology constraints (Cross-disciplinary Term Disambiguation, Intra-discipline Term Definitions, Symbols \& Constants Standardization, Variable Naming Consistency), and process constraints (Numerical Methods, Experimental Methods).
Constraints are format-agnostic; we do not require a rigid template.
A constraint passes only when the model states explicit, problem-grounded evidence tied to the instance’s symbols and values, rather than generic boilerplate.

\subsection{Construction Pipeline}
\label{sec:construction_qc}

We construct candidate instances and apply a three-stage quality control pipeline.
The stages are designed to separate three failure modes: ill-posed problems, incorrect references, and ungrounded constraints.
Only instances that pass all stages are included in the final benchmark.

\paragraph{Stage 1: Problem Validity.}
Goal: ensure the problem statement itself is well-posed.
We check that the prompt provides sufficient information to determine the target quantity, that the scientific setting is internally consistent, and that the enabled constraints are relevant to the task described by the prompt.
Instances fail this stage if the question is under-specified, contradictory, or if a constraint is enabled without any anchor in the problem statement that makes it applicable.

\paragraph{Stage 2: Reference Correctness.}
Goal: ensure the reference solution provides the correct scientific outcome.
Independently of constraint evidence, we solve or verify the problem and confirm that $y^\star$ matches the requested target result and key conclusions.
Instances fail this stage if the reference contains numerical errors, incorrect reasoning that changes the result, or mismatches the stated targets.

\paragraph{Stage 3: Constraint Grounding and Auditability.}
Goal: ensure constraints are truly binding and auditable for this instance.
We verify that every enabled constraint is instantiated by the prompt and that the reference explicitly provides the required evidence to satisfy that constraint.
This stage is necessary because correctness alone does not guarantee that a constraint is meaningful or checkable.
Instances fail this stage when a constraint is nominal rather than binding, or when the reference does not contain explicit, problem-grounded evidence that would allow a judge to audit compliance.

\paragraph{Stage 4: Domain Expert Review.}
Goal: ensure the instance reflects realistic scientific practice and remains interpretable to practitioners.
After passing automated and internal checks, each candidate is reviewed by a domain expert who assesses (i) whether the scientific context and assumptions are plausible for the stated domain, (ii) whether quantities, units, and parameter ranges are scientifically meaningful, and (iii) whether the reference solution communicates conclusions in a way consistent with field conventions.
Instances fail this stage if an expert flags the setting as implausible, identifies domain-knowledge contradictions, or judges that the prompt or reference would mislead a practitioner despite being formally solvable.

\begin{figure}
\includegraphics[width=0.5\textwidth]{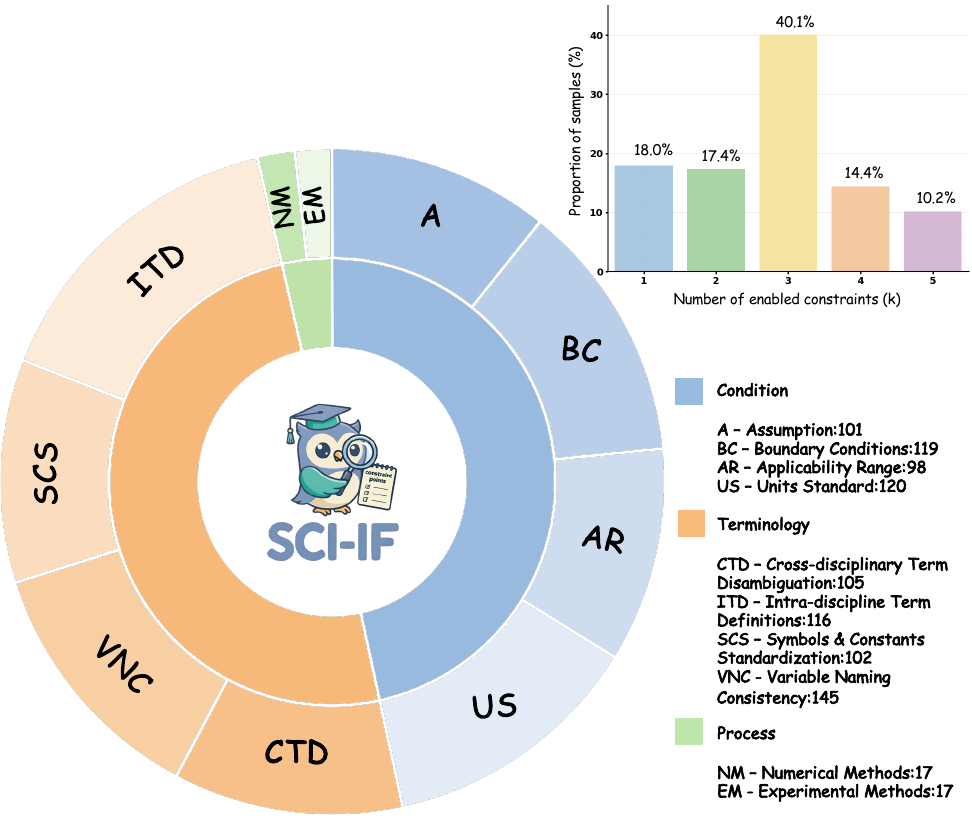}
\captionof{figure}{Constraint composition of \alias. Top-right: distribution of the number of enabled constraints per problem. Left: hierarchical breakdown of the constraint catalog, with inner wedges for the three constraint families and outer wedges for the 10 atomic constraints; wedge areas are proportional to how often each constraint is enabled in the test set.}
\label{fig:numconstraints}
\end{figure}

\begin{table*}[t]
  \centering
  \resizebox{\textwidth}{!}{
\begin{tabular}{l|cccc|c|cccc|c|cc|c}
    \toprule
    \textbf{Model} &
    \multicolumn{5}{c|}{\textbf{Condition}} &
    \multicolumn{5}{c|}{\textbf{Terminology}} &
    \multicolumn{3}{c}{\textbf{Process}} \\
    \midrule
    &
    \textbf{A} & \textbf{BC} & \textbf{AR} & \textbf{US} & \textbf{Avg.} &
    \textbf{CTD} & \textbf{ITD} & \textbf{SCS} & \textbf{VNC} & \textbf{Avg.} &
    \textbf{NM} & \textbf{EM} & \textbf{Avg.} \\
    \midrule

    \multicolumn{14}{l}{\textbf{Closed-source models}} \\
    \midrule
    GPT-5.2~\citep{openai_gpt52_2025}       & 65.3\% & \textbf{74.8\%} & \textbf{32.7\%} & 38.3\% & 53.2\% & 52.4\% & \textbf{81.9\%} & 41.2\% & \textbf{73.1\%} & \textbf{63.7\%} & 82.4\% & 94.1\% & \textbf{88.3\%} \\
    GPT-5.1~\citep{openai_gpt51_2025}       & \textbf{71.3\%} & 63.0\% & 30.6\% & \textbf{45.8\%} & \textbf{53.4\%} & 54.3\% & 74.1\% & 47.1\% & 72.4\% & 62.8\% & 70.6\% & 88.2\% & 79.4\% \\
    GPT-4o~\citep{openai_gpt4o_2024}        &  1.0\% &  0.0\% &  2.0\% &  0.0\% & 0.7\% &  2.9\% &  0.0\% &  4.9\% & 10.3\% & 4.8\% &  0.0\% &  0.0\% & 0.0\% \\
    GPT-o3~\citep{openai_o3_2025}        & 36.6\% & 46.2\% & 15.3\% & 19.2\% & 30.0\% & 29.5\% & 41.4\% & 27.5\% & 38.6\% & 34.7\% & 58.8\% & 76.5\% & 67.7\% \\
    GPT-o4mini~\citep{openai_gpt4omini_2024}    & 42.6\% & 38.7\% & 12.2\% & 19.2\% & 28.6\% & 28.6\% & 25.0\% & 18.6\% & 40.0\% & 29.0\% & 76.5\% & 82.4\% & 79.5\% \\
    Gemini-3~\citep{google_gemini3_2025}      & 52.5\% & 52.9\% & 23.5\% & 20.8\% & 38.0\% & 53.8\% & 65.5\% & 47.1\% & 69.0\% & 59.7\% & \textbf{88.2\%} & 82.4\% & 85.3\% \\
    Grok-4~\citep{xai_grok4_2025}        & 43.6\% & 53.8\% & 15.3\% & 33.3\% & 37.3\% & \textbf{70.5\%} & 59.5\% & \textbf{53.9\%} & 60.7\% & 61.0\% & 58.8\% & \textbf{100.0\%} & 79.4\% \\
    Claude-4.5Sonnet~\citep{Anthropic_Claude_Sonnet_4_5}
                 & 39.6\% & 47.1\% & 26.5\% & 25.8\% & 35.0\% & 36.2\% & 42.2\% & 35.3\% & 49.0\% & 41.5\% & 52.9\% & 82.4\% & 67.7\% \\
    Qwen3-Max~\citep{Qwen3_Max_Official_Blog}     & 40.6\% & 47.9\% & 15.3\% & 18.3\% & 31.0\% & 25.7\% & 44.0\% & 21.6\% & 55.2\% & 38.0\% & 82.4\% & 64.7\% & 73.6\% \\

    \midrule

    \multicolumn{14}{l}{\textbf{Open-source models}} \\
    \midrule

    Minimax-M2 ~\citep{MiniMax_M2_Series_Official}      & 18.8\% & 33.6\% &  5.1\% & 18.3\% & 19.5\% & 15.2\% & 16.4\% &  9.8\% & 33.1\% & 19.8\% & 52.9\% & 58.8\% & 55.9\% \\
    Kimi-K2 ~\citep{Kimi_K2_ArXiv_2025}      & 31.7\% & \textbf{37.8\%} & \textbf{17.3\%} & \textbf{23.3\%} & \textbf{27.8\%} & 25.7\% & 31.9\% & \textbf{26.5\%} & 45.5\% & 33.4\% & \textbf{88.2\%} & 58.8\% & 73.5\% \\
    GLM-4.7  ~\citep{ZAI_GLM_4_7_Docs}         & \textbf{41.6\%} & 36.1\% & \textbf{17.3\%} & 15.8\% & \textbf{27.8\%} & \textbf{43.8\%} & \textbf{44.8\%} & \textbf{26.5\%} & \textbf{53.1\%} & \textbf{42.8\%} & 58.8\% & \textbf{100.0\%} & \textbf{79.4\%} \\
    Qwen3-235b ~\citep{Qwen3_Tech_Report_2025}   & 24.8\% & 35.3\% &  9.2\% & 17.5\% & 22.1\% & 15.2\% & 20.7\% & 15.7\% & 31.7\% & 21.6\% & 76.5\% & 64.7\% & 70.6\% \\
    Qwen3-80b  ~\citep{Qwen3_Tech_Report_2025}   & 25.7\% & 36.1\% & 11.2\% &  8.3\% & 20.5\% & 17.1\% & 10.3\% & 24.5\% & 31.7\% & 21.2\% & 35.3\% & 58.8\% & 47.1\% \\
    Deepseek-v3.2 ~\citep{DeepSeek_V3_2_Official_Overview}& 31.7\% & 34.5\% & 12.2\% & 11.7\% & 22.8\% & 17.1\% & 21.6\% & 20.6\% & 34.5\% & 24.0\% & 64.7\% & 64.7\% & 64.7\% \\
    \bottomrule
  \end{tabular}
  }
  \caption{\label{tab:per_constraint_strict_abbrev}
    Per-constraint correctness under the Strict mode. Abbrev.:
    A=Assumptions; BC=Boundary Conditions; AR=Applicability Range; US=Units Standard;
    CTD=Cross-disciplinary Term Disambiguation; ITD=Intra-discipline Term Definitions;
    SCS=Symbols \& Constants Standardization; VNC=Variable Naming Consistency;
    NM=Numerical Methods; EM=Experimental Methods.
  }
\end{table*}

\subsection{Evaluation and Auditing}
\label{sec:evaluation_pipeline}

\alias\ follows a \emph{generate--then--audit} protocol. Given a problem and its enabled constraint set, the model first produces an answer $\hat{y}$. We then audit \textbf{answer correctness} and \textbf{constraint compliance} as two separate axes, so that being numerically correct can be distinguished from being scientifically auditable.

\paragraph{Constraint Auditing and Aggregation.}
For each enabled constraint $c \in C$, the auditor assigns PASS or FAIL by checking whether the model output contains \textbf{explicit, instance-grounded evidence} for that constraint. Evidence must be clearly stated and tied to the concrete symbols and values in the problem. If evidence is missing, vague, or contradicted by the solution, the constraint fails. For items with multiple constraints, we aggregate with a logical AND over enabled constraints: an item is compliant only if every $c \in C$ passes. This makes failures directly attributable at the constraint level.

\paragraph{Correctness as an Independent Axis.}
Correctness is audited independently by comparing $\hat{y}$ to the reference $y^\star$ under task-appropriate equivalence criteria for the final outcome and key conclusions. The correctness audit ignores constraint-following issues to avoid conflating scientific reporting failures with arithmetic or reasoning errors. This separation isolates two common regimes in scientific QA: correct-but-non-compliant outputs and compliant-but-incorrect outputs.

\subsection{Hybrid Auditing}
\label{sec:dual_judge}

\paragraph{Rules First, Judges as Fallback.}
To maximize reproducibility and minimize subjectivity, we use a hybrid auditing pipeline that prioritizes deterministic checks. We first apply a rule-based auditor for signals that admit stable verification, including parsing LaTeX math expressions, checking numerical equivalence under tolerances, validating required unit declarations, and matching constraint-specific evidence patterns. When a rule-based check is not applicable or yields an unreliable decision, we fall back to model-based judging under a strict evidence standard.

\paragraph{Dual Judging with Aligned Standards.}
Model-based judging uses two independent judges, GPT-5.1 and Gemini-3-Flash, with a conservative agreement rule: a constraint passes only if both judges mark it PASS. In practice, different judges can apply systematically different strictness even on the same response. We therefore align judging standards using a held-out calibration set that is never used for reported results. We identify recurring disagreement modes and encode the corresponding decision thresholds into a fixed, standardized judging prompt. The standardized prompt instructs judges to evaluate evidence presence and consistency only, and to avoid reconstructing missing checks, assumptions, or derivations. After calibration, the judging prompts are fixed for all experiments.

%% file: section/experiments.tex
\section{Experiments}
\label{sec:experiments}

We evaluate whether scientific QA models can be \emph{both} correct and scientifically auditable. \alias\ reports two metrics per model: \textbf{answer correctness} and \textbf{multi-constraint compliance}.

\subsection{Setup}
\label{sec:exp_setup}

We evaluate on the 334-item \alias\ test set spanning Biology, Chemistry, Materials, and Physics. We include a mix of frontier and open models, with the full list deferred to Appendix~\ref{app:exp_models}. All models use identical decoding settings and the same prompting template (Appendix~\ref{app:eval_prompts}). Compliance and correctness are scored with our dual-judge auditing protocol.

\paragraph{Post-Training Variants.}
To test whether SciIF training improves both scientific QA and scientific auditability, we post-train Qwen3-8B on our 910-item training set in two stages. \textbf{SFT} targets stronger scientific problem solving and higher answer correctness. We then apply \textbf{verifier-based RL} that rewards explicit, problem-grounded constraint evidence, encouraging the model to proactively surface domain-relevant validity information such as assumptions, applicability limits, unit discipline, and boundary checks. We evaluate the base model, \alias-SFT, and \alias-RL on \alias\ and on external benchmarks (Appendix~\ref{app:posttraining}).

\begin{table*}[t]
  \centering
  \resizebox{\textwidth}{!}{
  \begin{tabular}{lccccccc}
\toprule
Model & Compliance$\uparrow$ & Correctness$\uparrow$ & IFEval (Strict)$\uparrow$ & MMLU-Bio$\uparrow$ & MMLU-Chem$\uparrow$ & MMLU-Phys$\uparrow$ & MMLU Avg$\uparrow$ \\
\midrule
qwen3-8B (Base) & 2.1\% & 24.3\% & 80.4\% & 61.0\% & \textbf{61.0\%} & 68.0\% & 63.3\% \\
w/ IFEval~(SFT) & --- & --- & --- & 43.0\% (-18.0\%) & 37.0\% (-24.0\%) & 61.0\% (-7.0\%) & 47.0\% (-16.3\%) \\
w/ \alias~(SFT) & 3.1\% (+1.0\%) & 30.2\% (+5.9\%) & 81.7\% (+1.3\%) & \textbf{67.0\%} (+6.0\%) & 58.0\% (-3.0\%) & \textbf{76.0\%} (+8.0\%) & \textbf{67.0\%} (+3.7\%) \\
w/ \alias~(SFT+RL) & \textbf{7.0\%} (+4.9\%) & \textbf{31.7\%} (+7.4\%) & \textbf{83.2\%} (+2.8\%) & --- & --- & --- & --- \\
\bottomrule
\end{tabular}
}
\caption{Transfer effects beyond \alias. IFEval (Strict) reports pass rate. MMLU reports accuracy on three STEM subjects and their macro-average.}
\label{tab:transfer_if_mmlu}
\end{table*}

\subsection{Correctness is not Compliance}
\label{sec:exp_gap}

Our central finding is a persistent gap between getting the right answer and meeting scientific constraints. Across all evaluated models, \textbf{answer correctness is substantially higher than strict multi-constraint compliance}. Even strong models exceed $80\%$ correctness, yet overall multi-constraint pass remains below $30\%$ (best: $29.6\%$). This pattern holds across model families, indicating a systematic failure mode: models often reach the correct final result while omitting explicit evidence for one or more constraints that govern validity, meaning, and reproducibility.

Among closed-source models, GPT-5.2 and Grok-4 demonstrate superior performance, excelling in language understanding and computational reasoning respectively. In the open-source category, GLM-4.7 shows particularly strong capabilities across most dimensions.

\begin{figure*}
    \centering
    \vspace{-5pt}
    \includegraphics[width=0.78\linewidth]{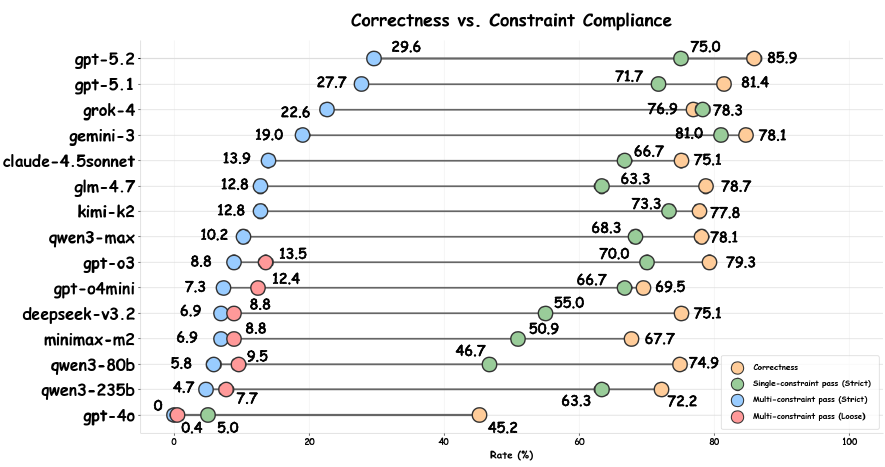}
    \vspace{-5pt}
    \caption{Comparison of three metrics: answer correctness, single-constraint pass, and multi-constraint overall pass. Single-constraint pass is computed on the 60 items with exactly one enabled constraint. Multi-constraint overall pass is computed on the 274 items with multiple enabled constraints, where an item passes only if all enabled constraints pass.}
    \label{fig:compliance}
\end{figure*}

We furthre visualize this gap in Figure~\ref{fig:compliance}.
For most models, single-constraint pass is relatively strong, which indicates that many scientific requirements are individually achievable when they are the only thing to remember. The failure appears when constraints must be coordinated: multi-constraint overall pass collapses to a much lower regime even when correctness stays high. This separation shows that the dominant error is not inability to solve the underlying scientific problem, but inability to reliably produce a solution as a reproducible artifact with explicit, problem-grounded scientific discipline. In practice, this means an accuracy-only evaluation can label an output as “solved” even when it omits units, leaves assumptions implicit, fails to state validity limits, or drifts in symbol meaning—issues that can silently break interpretation or downstream reuse.

\subsection{Compositional Effects}
\label{sec:exp_joint_comp}

Correctness and compliance are related but asymmetric. Compliant answers are typically correct, while correct answers are frequently non-compliant, so accuracy-only evaluation overestimates scientific instruction-following ability. We also observe a consistent \textbf{compositional collapse}: compliance drops sharply as the number of enabled constraints increases from $k{=}2$ to $k{=}5$ (Appendix~\ref{app:pass_vs_k}), revealing a coordination bottleneck in maintaining multiple scientific requirements within one derivation.

\subsection{Generalizability}
\label{sec:exp_transfer}

We test whether training on SciIF data improves performance outside \alias. Starting from Qwen3-8B, we compare the base model to an \alias-SFT checkpoint and an \alias-RL checkpoint. Table~\ref{tab:transfer_if_mmlu} shows consistent gains on IFEval (Strict) (Appendix~\ref{app:ifeval_check}) and improved performance on MMLU science subjects (Appendix~\ref{app:mmlu_case_study_scif} Fig.~\ref{fig:mmlu_case_study_physics0119}), suggesting that constraint-oriented post-training can transfer to broader instruction adherence and scientific QA rather than only improving in-domain compliance (Appendix~\ref{app:eval_protocol}).

\subsection{Where Models Break}
\label{sec:exp_where_break}

Constraint-level analysis shows that failures concentrate on global semantic requirements such as Units Standard and Applicability Range, which demand explicit, problem-grounded statements and penalize silent assumptions. Full per-constraint breakdowns are reported in Appendix~\ref{app:experiments}.

\subsection{Explicit rubrics reduce judge variance}
\label{sec:exp_prompt_rubric}

Because our evaluation uses LLM-as-a-judge, the judge prompt effectively defines the scoring rule.
When constraints are under-specified, different judge models apply different implicit thresholds (e.g., treating $0.1$ vs.\ $0.01$ error as ``correct''), leading to non-comparable compliance scores.
We therefore make scoring points and decision thresholds explicit in the prompt (Appendix~\ref{app:judge_prompts}).
This calibration sharply reduces inter-judge discrepancy: before clarification, GPT-5.1 vs.\ Gemini-3-Flash yields $68.7\%$ vs.\ $88.0\%$; after clarification, $79.1\%$ vs.\ $81.4\%$, reducing the gap from 19.3 to 2.3 points. This motivates treating prompt-level rubric specification as a first-class part of protocol.

\subsection{Compositional Collapse: Patterns and Anomalies across Models}
\label{sec:compositional_collapse_main}

\begin{figure}[t]
  \centering
  \includegraphics[width=0.5\textwidth]{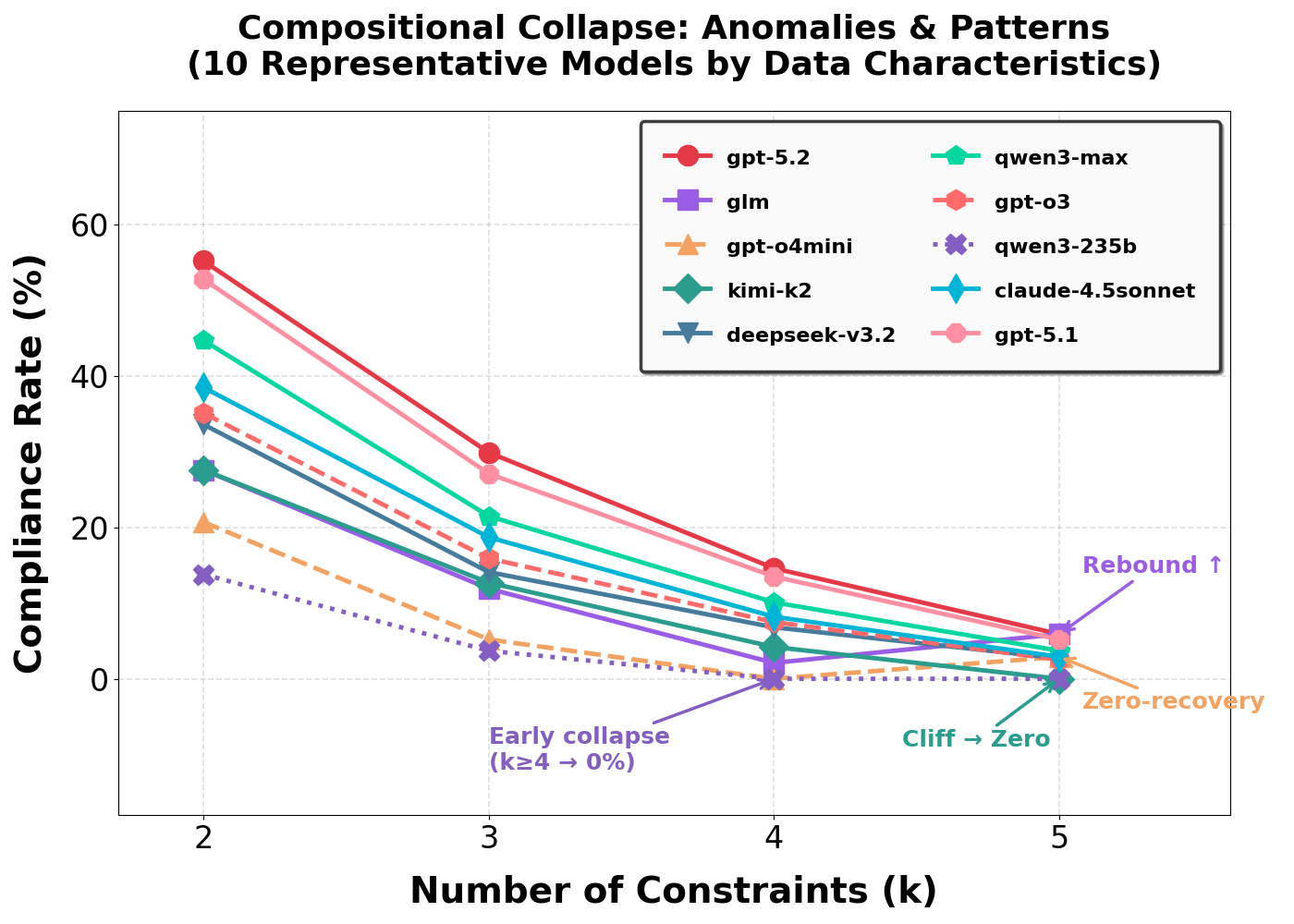}
  \caption{Compositional collapse under increasing constraint load: strict compliance rate versus the number of enabled constraints $k$ for ten representative models.}

% Most models exhibit steep, near-monotonic decay as $k$ increases, consistent with a global constraint-coordination bottleneck.
% A few non-monotonic ``rebound'' patterns appear at high $k$, which we treat as diagnostic artifacts of rubric thresholding and small-sample variance rather than robust gains in compositional ability.}
  \label{fig:collapse_patterns}
\end{figure}

Beyond aggregate scores, the per-$k$ breakdown reveals a consistent \emph{coordination bottleneck}.
As the number of enabled constraints increases, compliance drops steeply for nearly all models,
even when answer correctness remains comparatively stable.
Figure~\ref{fig:collapse_patterns} visualizes this effect for ten representative models selected to cover
typical and atypical curve shapes.

\paragraph{A shared baseline shape: smooth exponential-like decay.}
Strong closed models such as GPT-5.2 and GPT-5.1 exhibit a similar high-start curve at $k{=}2$,
followed by a near-monotonic decline through $k{=}5$.
This pattern suggests that failures are rarely caused by a single missing skill.
Instead, they arise from \emph{global constraint scheduling}: maintaining unit discipline, symbol meaning,
and required validity checks while completing the derivation.

\paragraph{Cliff-drop regimes: fragile constraint composition in open models.}
Several models enter a ``cliff'' regime where compliance collapses rapidly once $k{\ge}3$.
For example, Qwen3-235b reaches low but non-zero compliance at $k{=}2$ and $k{=}3$,
then drops to $0\%$ for $k{\ge}4$.
This shape indicates that the model can sometimes satisfy isolated constraints,
but fails to keep multiple constraints active across a longer solution trajectory.

\paragraph{Non-monotonic anomalies: brittleness and measurement effects.}
A small set of models exhibit non-monotonic behavior, including partial rebounds at $k{=}5$,
or ``zero-then-recovery'' patterns.
We treat these as diagnostics rather than evidence of improved compositional ability.
Two factors can produce such anomalies:
(i) small-sample variance at high $k$ due to fewer items,
and (ii) discrete rubric thresholds where a model occasionally ``remembers'' to add a missing evidence sentence
that flips an AND-aggregated decision from FAIL to PASS.
These anomalies reinforce a key point: compliance failures are often driven by \emph{missing explicit evidence},
not necessarily by incorrect scientific reasoning.

\subsection{Takeaway}
\label{sec:exp_summary}

\alias\ makes visible a gap that correctness-centric benchmarks systematically miss: \textbf{scientific correctness and scientific auditability are distinct capabilities}. Current models often produce correct results without reliably producing the explicit evidence needed for reproducible scientific reasoning, and this weakness becomes sharper under multi-constraint composition.

%% file: section/appendix.tex
\clearpage

\phantomsection
\apptocheading{Appendix Contents}
\addcontentsline{toc}{section}{Appendix Contents}

% ---- Manual entries (edit as your appendix evolves) ----
\apptocsection{app:eval_prompts}{A\quad Evaluation Protocol and Prompt Templates}
\apptocsubsection{app:eval_protocol}{A.1\quad Generate--then--audit protocol}
\apptocsubsection{app:verifier_pipeline}{A.2\quad Rule-based verifier with judge fallback}
\apptocsubsection{app:gen_prompts}{A.3\quad Answer generation prompts}
\apptocsubsection{app:judge_prompts}{A.4\quad Judge prompts}
\apptocsubsection{app:decision_policies}{A.5\quad Decision policies}
\apptocsubsection{app:strict_loose_summary}{A.6\quad Strict vs.\ Loose summary}
\apptocsubsection{app:ifeval_check}{A.7\quad External consistency check on instruction following}
\apptocsubsection{app:rubric_summary}{A.8\quad Constraint rubric summary}

\bigskip
\apptocsection{app:benchmark_details}{B\quad Benchmark Details}
\apptocsubsection{app:source_type_composition}{B.1\quad Source-type composition of the 1244-item pool}
\apptocsubsection{app:rubrics}{B.2\quad Evidence-based rubrics}
\apptocsubsection{app:regen}{B.3\quad Optional regeneration and stability filtering}
\apptocsubsection{app:judge_calib}{B.4\quad Judge scale alignment}
\apptocsubsection{app:judge_case_study}{B.5\quad Case study: judge disagreement and prompt calibration}

\bigskip
\apptocsection{app:posttraining}{C\quad Post-Training Details}
\apptocsubsection{app:ifeval_case_study}{C.1\quad Case study: what verifier-based RL changes on IFEval}
\apptocsubsection{app:mmlu_case_study_scif}{C.2\quad Case study: SciIF SFT improves structured reasoning on MMLU}

\bigskip
\apptocsection{app:experiments}{D\quad Additional Experimental Results and Diagnostics}
\apptocsubsection{app:exp_models}{D.1\quad Models and inference settings}
\apptocsubsection{app:pass_vs_k}{D.2\quad Compositional collapse vs.\ number of constraints}

\apptocsection{app:ai_usage}{E\quad AI Assistant Use}
\apptocsection{app:human_protocol}{F\quad Human Validation Protocol for Equivalence Between Model and Reference Answers}

\section{Evaluation Protocol and Prompt Templates}

\label{app:eval_prompts}

This appendix provides the information needed to reproduce our evaluation and training signals.
We first document the generate--then--audit protocol and the rule-based \mbox{verifier--judge} pipeline,
then list the answer-generation prompts, judge prompts, and decision policies used in Strict and Loose modes.

\subsection{Generate--then--audit protocol}
\label{app:eval_protocol}

\paragraph{Two-axis evaluation.}
Each model output is evaluated on two axes:
answer correctness and constraint compliance.
Correctness checks whether the required scientific outcome matches the reference under task-appropriate equivalence criteria.
Compliance checks whether the output contains explicit, problem-grounded evidence for each enabled constraint.
Judges evaluate evidence presence rather than infer missing checks, assumptions, or definitions.

\paragraph{Per-constraint auditing and aggregation.}
Each enabled constraint is audited independently and receives PASS/FAIL.
For multi-constraint items, item-level compliance is the logical AND across enabled constraints.

\paragraph{Dual-judge robustness.}
We use two independent judges and apply a two-of-two rule.
A constraint passes only if both judges return PASS, reducing sensitivity to single-judge drift and stylistic preference.

\subsection{Rule-based verifier with judge fallback}
\label{app:verifier_pipeline}

We prioritize deterministic checks when feasible and fall back to LLM judges when a rule-based check
is not applicable or fails to parse the output.
Rule-based checks include numerical equivalence under task-defined criteria, expression parsing when needed,
unit and dimensional consistency checks when units are stated, and mechanical substitution checks for boundary or limiting cases.
When a rule cannot be executed reliably, the decision is delegated to the judge prompts below.

\subsection{Answer generation prompts}
\label{app:gen_prompts}

\begin{promptbox}{A.1 Strict Mode: Answer Generation Prompt}
You are solving a university-level scientific problem.

PROBLEM (verbatim):
[RAW_QUESTION_WITH_CONSTRAINTS]

Enabled constraints:
[CONSTRAINT_NAME_LIST]

Required evidence (MUST appear; missing any item => non-compliant):
[CONSTRAINT_SPECIFIC_EVIDENCE_POINTS]

Writing rules:
- Integrate evidence into the solution narrative with natural transitions.
- Do not use checklist-style rubric blocks or fixed labels.
- Do not claim compliance with slogans; show explicit, auditable statements and checks.
- Do not use markdown tables for definitions; define symbols inline in sentences.
- Follow the problem's stated units and precision requirements.

Answer format:
- Short plan
- Derivation / reasoning (equations if needed)
- Final Answer (clearly stated, with units if applicable)

Answer:
\end{promptbox}

\subsection{Judge prompts}
\label{app:judge_prompts}

We audit correctness and constraint compliance separately.
For compliance, judges evaluate one enabled constraint at a time against explicit evidence points and return a pointwise verdict.

\begin{judgebox}{A.3 Strict Mode: Per-Constraint Judge Prompt}
You are an expert validator for university-level [SUBJECT] problems.

Constraint name:
[CONSTRAINT_NAME]

Constraint description:
[CONSTRAINT_DESCRIPTION]

Required evidence points (evaluate EACH independently):
POINT_1: [POINT_1_DESC]
POINT_2: [POINT_2_DESC]
...

Problem:
[QUESTION]

Reference answer (if provided):
[GOLD_ANSWER]

Model answer:
[MODEL_ANSWER]

Decision policy:
- Decide whether each point has explicit, problem-grounded evidence in the model answer.
- Do NOT infer missing evidence.
- If the model claims a check, verify it is correct and consistent with its own work.

Output format (MUST follow exactly):
POINT_1: YES or NO [brief reason if NO]
POINT_2: YES or NO [brief reason if NO]
...
OVERALL: YES or NO
OVERALL_REASON: [<=150 chars, list failed points]
\end{judgebox}

\begin{judgebox}{A.4 Loose Mode: Per-Constraint Judge Prompt}
You are an expert validator for university-level [SUBJECT] problems.

Constraint name:
[CONSTRAINT_NAME]

Constraint description:
[CONSTRAINT_DESCRIPTION]

Required evidence points (evaluate EACH independently):
POINT_1 [MAIN]: [POINT_1_DESC]
POINT_2 [MAIN]: [POINT_2_DESC]
POINT_3 [SECONDARY]: [POINT_3_DESC]
...

Problem:
[QUESTION]

Reference answer (if provided):
[GOLD_ANSWER]

Model answer:
[MODEL_ANSWER]

Decision policy:
- Same as STRICT for pointwise evidence, non-inference, and truthfulness checks.
- OVERALL is YES only if all MAIN points are YES. SECONDARY points may be NO.

Output format (MUST follow exactly):
POINT_1: YES or NO [brief reason if NO]
POINT_2: YES or NO [brief reason if NO]
POINT_3: YES or NO [brief reason if NO]
...
OVERALL: YES or NO
OVERALL_REASON: [<=150 chars, list failed MAIN points]
\end{judgebox}

\subsection{Decision policies}
\label{app:decision_policies}

\paragraph{Strict policy.}
A point is YES only when the model output contains explicit evidence that is problem-grounded, correct,
and consistent with the model's own derivation.
Missing evidence is NO.
Incorrect claimed checks are NO.

\paragraph{Loose policy.}
Loose mode keeps the same non-inference and truthfulness requirements, but only MAIN points
are required for overall PASS. SECONDARY points may fail without failing the constraint.

\subsection{Strict vs.\ Loose summary}
\label{app:strict_loose_summary}

We use two compliance modes to separate missing-evidence failures from substantive violations.
Strict is the primary reporting setting and is designed for auditability.
Loose is a diagnostic setting used to test whether low scores are driven mainly by omission of write-ups.

% -----------------------------
\section{Benchmark Details}
\label{app:benchmark_details}

This section provides additional information about rubric structure, instance construction, 
and judge calibration.We provide a high-level overview of the ten constraints in Table~\ref{tab:rubric_highlevel} to clarify what constitutes auditable evidence at a glance.The full point-wise definitions, edge cases, and the Strict/Loose point split are provided in the supplementary rubric specification.

\subsection{Source-type composition of the 1244-item pool}
\label{app:source_type_composition}

Our full data pool contains 1244 items and is intentionally dominated by textbook-style problems.
We use web-derived items as a secondary source to broaden coverage, while keeping QA-style items as a smaller component.
Overall, the source-type mixture follows an approximate 7:2:1 ratio:
about 70\% textbook, about 20\% web, and about 10\% QA.
This design emphasizes academically grounded problem structures while still injecting topical diversity and alternative phrasing patterns.

\begin{table}[t]
\centering
\small
\begin{tabular}{p{0.22\linewidth} p{0.34\linewidth} p{0.34\linewidth}}
\hline
\textbf{Aspect} & \textbf{Strict} & \textbf{Loose} \\
\hline
Overall verdict & All points must pass & All main points must pass \\
Missing evidence & Always FAIL & Always FAIL \\
Semantic checks & Applied broadly & Focused on main points \\
Use case & Main leaderboard, auditable scoring & Diagnosis of omission vs.\ violation \\
\hline
\end{tabular}
\caption{High-level differences between Strict and Loose compliance modes.}
\label{tab:strict_loose_summary}
\end{table}

\begin{table*}[t]
\centering
\small
\begin{tabular}{p{0.22\linewidth} p{0.18\linewidth} p{0.52\linewidth}}
\hline
\textbf{Constraint} & \textbf{Family} & \textbf{Main evidence required (high level)} \\
\hline
Assumptions & Condition & State at least one key assumption used, explain its impact on the method, and anchor it to a concrete step in this solution. \\
Boundary Conditions & Condition & Extract the boundary or hard conditions from the prompt and provide at least one mechanical verification that the derived solution satisfies them. \\
Applicability Range & Condition & Name the approximation or model used, state an explicit validity range, and describe a failure mode outside the range with a correct direction or trend. \\
Units Standard & Condition & Give units for key variables at first use, keep units consistent, and include a short dimensional or unit self-check tied to the target quantity. \\
Cross-disciplinary Disambiguation & Terminology & Identify ambiguous terms, state the intended meaning and a non-intended meaning, and point to where the intended meaning is used in the solution. \\
Intra-discipline Definitions & Terminology & Provide a plain definition and a formal criterion, then apply the criterion in one concrete step of the solution. \\
Symbols \& Constants Standardization & Terminology & Define symbols used in the final result, declare constants with units and source when required, and avoid symbol drift. \\
Variable Naming Consistency & Terminology & Maintain one symbol per quantity throughout the solution and prevent semantic drift in symbol meaning and units. \\
Numerical Methods & Process & Name the algorithm, provide the update rule using problem symbols, and show at least one instantiated numerical iteration. \\
Experimental Methods & Process & Provide an executable procedure tied to problem variables, include auditable anchors, and state how uncertainty affects the target quantity. \\
\hline
\end{tabular}
\caption{High-level rubric summary for the ten constraints. Detailed point-wise definitions and decision policies are in Appendix~A.3--A.4.}
\label{tab:rubric_highlevel}
\end{table*}

\subsection{Evidence-based rubrics}
\label{app:rubrics}

Each enabled constraint is accompanied by evidence points designed to be checkable without reconstruction or guesswork.
A constraint passes only if all required points pass.
Any missing, vague, or contradictory evidence causes failure.
We release the full point-level rubric specification in machine-readable form as supplementary material.

\subsection{Optional regeneration and stability filtering}
\label{app:regen}

For a small subset of items, generation can yield good problems but brittle references.
When needed, we regenerate reference candidates and retain instances that fall within a target stability window,
filtering out instances that are effectively trivial or near-impossible under the same constraints.

\subsection{Judge scale alignment}
\label{app:judge_calib}

Two-of-two voting reduces noise but does not guarantee that judges apply identical evidence thresholds.
We align judge scales using a held-out calibration set that is never used for reported scores.
We compare judge outputs per evidence point, identify recurring disagreement patterns,
and refine judge prompts by making those thresholds explicit and executable.
After calibration, prompts are fixed for all experiments.

% ---------------------------------------------

% ====== Case study box (two-column wide) ======
\subsection{Case study: judge disagreement and prompt calibration}
\label{app:judge_case_study}

We conducted a targeted case study to quantify and reduce judge arbitrariness.
We sampled 50 instances and re-audited the same model outputs with two judge models,
then compared pointwise PASS/FAIL decisions under the same rubric.
We observed non-trivial disagreement: 19 out of 50 instances had at least one constraint-level mismatch,
with 20 constraint-level mismatches in total.
Disagreements concentrated on globally semantic constraints such as
\textsc{Symbols \& Constants Standardization} and \textsc{Intra-discipline Definitions},
where the effective threshold for ``explicit evidence'' differed across judges.
These findings motivated prompt calibration that removes packaging-dependent expectations
and rewrites evidence points into problem-grounded, executable checks,
leaving minimal room for stylistic interpretation.

\begin{figure}[t]
    \centering
    \includegraphics[width=\linewidth]{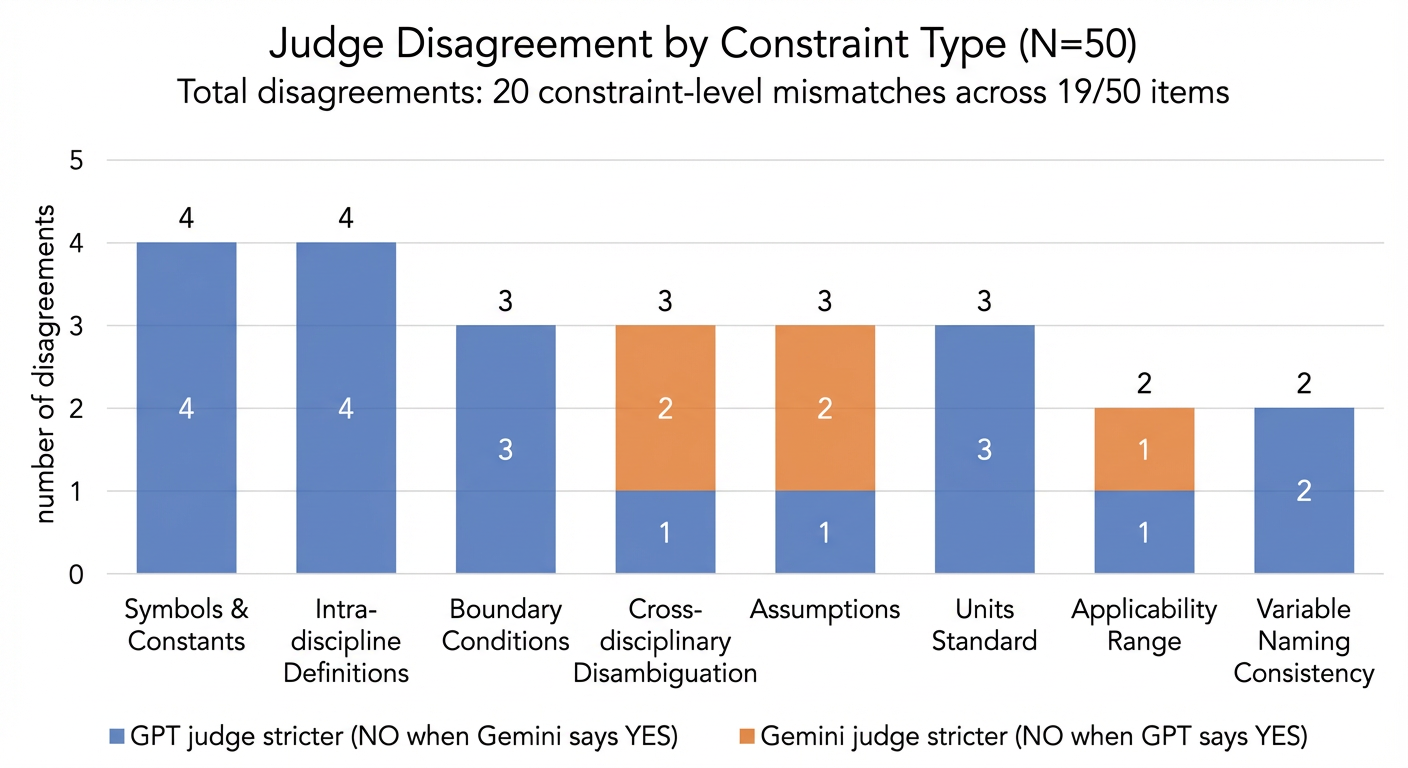}
    \caption{Judge disagreement by constraint type on a 50-item audit set.
Bars count constraint-level mismatches between two judges, split by which judge is stricter.
Across the 50 items, we observe 20 constraint-level mismatches spanning 19 items,
with disagreements concentrated in \textsc{Symbols \& Constants} and \textsc{Intra-discipline Definitions}.}
    \label{fig:judge_disagreement_by_constraint}
\end{figure}

\begin{figure*}[t]
\centering

\begin{tcolorbox}[
  enhanced,
  breakable,
  width=\textwidth,
  colback=quanback,
  colframe=quanframe,
  boxrule=0.8pt,
  arc=2mm,
  left=2.5mm,right=2.5mm,top=1.8mm,bottom=1.8mm,
  title={Pre-calibration ambiguity example: Biology-48, Symbols \& Constants Standardization},
  fonttitle=\bfseries\large,
  coltitle=black
]
\sloppy
\setlength{\parindent}{0pt}
\small

\textbf{Observed mismatch.}
Two judges disagreed on whether the model output satisfies \textsc{Symbols \& Constants Standardization}.
Both judges accepted the computations, but they differed on whether $S_1$ and $S_2$
must be explicitly declared with units rather than treated as instances of a generic substrate symbol.

\vspace{0.6em}

% ---------- Two-column content, robust against overflow ----------
\begin{tcbraster}[
  raster columns=2,
  raster column skip=3mm,
  raster equal height=rows,
  raster left skip=0pt,
  raster right skip=0pt
]
\begin{tcolorbox}[
  enhanced,
  colback=white,
  colframe=quanframe!45,
  boxrule=0.6pt,
  arc=1.5mm,
  left=2mm,right=2mm,top=1.2mm,bottom=1.2mm,
  title={Problem (verbatim excerpt)},
  fonttitle=\bfseries
]
\ttfamily\scriptsize
A purified eukaryotic enzyme follows Michaelis--Menten kinetics in a 1.00 L assay at $37^\circ$C.\\
Measured parameters:\\
$V_{\max}=1.20\times10^{-4}\ \mathrm{mol\,L^{-1}\,min^{-1}}$, \ 
$K_m=5.0\times10^{-5}\ \mathrm{mol\,L^{-1}}$, \ 
$E_t=2.0\times10^{-6}\ \mathrm{mol\,L^{-1}}$.\\
Substrate concentrations:\\
$S_1=2.5\times10^{-5}\ \mathrm{mol\,L^{-1}}$, \ 
$S_2=2.0\times10^{-4}\ \mathrm{mol\,L^{-1}}$.\\
Compute:\\
(1) $k_{\mathrm{cat}}$ in $\mathrm{s^{-1}}$;\\
(2) $v(S_1),v(S_2)$ in $\mathrm{mol\,L^{-1}\,min^{-1}}$;\\
(3) $k_{\mathrm{cat}}/K_m$ in $\mathrm{L\,mol^{-1}\,s^{-1}}$;\\
(4) Avogadro-based molecules-per-active-site-per-second check.
\end{tcolorbox}

\begin{tcolorbox}[
  enhanced,
  colback=white,
  colframe=quanframe!45,
  boxrule=0.6pt,
  arc=1.5mm,
  left=2mm,right=2mm,top=1.2mm,bottom=1.2mm,
  title={Model output (relevant excerpt)},
  fonttitle=\bfseries
]
\ttfamily\scriptsize
$v(S_1)=V_{\max}\cdot S_1/(K_m+S_1)$\\
\ \ $=(1.20\times10^{-4})(2.5\times10^{-5})/(5.0\times10^{-5}+2.5\times10^{-5})$\\
\ \ $=4.0\times10^{-5}\ \mathrm{mol\,L^{-1}\,min^{-1}}$.\\
$v(S_2)=V_{\max}\cdot S_2/(K_m+S_2)$\\
\ \ $=(1.20\times10^{-4})(2.0\times10^{-4})/(5.0\times10^{-5}+2.0\times10^{-4})$\\
\ \ $=9.6\times10^{-5}\ \mathrm{mol\,L^{-1}\,min^{-1}}$.\\[0.3em]
Symbols declared in final answer:\\
$v$ [$\mathrm{mol\,L^{-1}\,min^{-1}}$], $V_{\max}$ [$\mathrm{mol\,L^{-1}\,min^{-1}}$],\\
$K_m$ [$\mathrm{mol\,L^{-1}}$], $S$ [$\mathrm{mol\,L^{-1}}$], $E_t$ [$\mathrm{mol\,L^{-1}}$],\\
$k_{\mathrm{cat}}$ [$\mathrm{s^{-1}}$], $N_A$ [$\mathrm{mol^{-1}}$].\\
Constants: $N_A=6.02214076\times10^{23}\ \mathrm{mol^{-1}}$ (CODATA 2018/2022).\\
No drift: $v$ denotes velocity in $\mathrm{mol\,L^{-1}\,min^{-1}}$.
\end{tcolorbox}
\end{tcbraster}

\vspace{0.8em}
\textbf{Constraint under audit.} \textsc{Symbols \& Constants Standardization}

\vspace{0.3em}
\textbf{Evidence points (operationalization).}
\begin{itemize}
    \item \textbf{POINT\_1 Symbol declaration:} all symbols used in computation are explicitly declared with units at least once.
    \item \textbf{POINT\_2 Constant declaration:} any constant used provides numeric value, unit, and standard source when required.
    \item \textbf{POINT\_3 No drift:} symbols do not change meaning or units across the solution.
    
\end{itemize}

\vspace{0.5em}
\textbf{Pre-calibration judge outputs (pointwise).}

\renewcommand{\arraystretch}{1.15}
\begin{tabularx}{\textwidth}{@{}>{\raggedright\arraybackslash}p{0.27\textwidth} c c >{\raggedright\arraybackslash}X@{}}
\toprule
\textbf{Evidence point} & \textbf{Gemini} & \textbf{GPT} & \textbf{Reason for mismatch} \\
\midrule
POINT\_1 Symbol declaration & YES & NO &
GPT judge required an explicit declaration for $S_1$ and $S_2$ with units and did not accept
treating $S_1,S_2$ as implicitly covered by a generic $S$ declaration. \\
POINT\_2 Constant declaration & YES & YES &
Both judges accepted $N_A$ with value, unit, and CODATA source. \\
POINT\_3 No drift & YES & YES &
Both judges found no meaning or unit drift for declared symbols. \\
\bottomrule
\end{tabularx}

\vspace{0.7em}
\textbf{Calibration change applied.}
We revised the evidence wording to remove packaging dependence and eliminate judge-specific degrees of freedom.
The updated requirement is model-independent and executable: for every constant or symbol that appears in computation,
the answer must state its meaning and unit at least once, either inline or in a short declaration sentence.
We clarified that a dedicated symbol table is not required, and that a generic symbol declaration does not automatically
cover indexed instances unless the indexing relation is stated.

\end{tcolorbox}

\caption{A concrete pre-calibration disagreement case.
The same model output is accepted by one judge and rejected by the other due to different implicit thresholds for symbol explicitness.
Prompt and rubric calibration rewrites evidence points into packaging-neutral, executable requirements to reduce judge-specific interpretation space.}
\label{fig:judge_case_biology48}
\end{figure*}

\subsection{External consistency check on instruction following}
\label{app:ifeval_check}

As an external sanity check, we evaluated Qwen3-8B and its post-trained variants on IFEval under Strict and Loose settings.
Verifier-guided reinforcement learning improves instruction following beyond supervised fine-tuning alone, with the RL variant achieving higher Strict performance than both the base and SFT variants.
This supports the interpretation that constraint-oriented post-training strengthens general instruction discipline rather than only improving in-domain compliance on \alias.
% ---------------------------------------------

\subsection{Constraint rubric summary}
\label{app:rubric_summary}

We define ten atomic constraints grouped into three families.
Table~\ref{tab:rubric_highlevel} summarizes the main evidence requirements at a high level.

% (keep your existing Table~\ref{tab:rubric_highlevel} here unchanged)

% -----------------------------
\section{Post-Training Details}
\label{app:posttraining}

This section documents the SFT and verifier-based RL objectives, reward construction, and implementation notes.

\subsection{SFT objective}
We train a LoRA adapter on top of Qwen3-8B with the standard next-token likelihood objective:
\begin{equation}
\mathcal{L}_{\mathrm{SFT}}(\theta) \;=\; - \sum_{t} \log \pi_\theta\!\big(y_t^\ast \mid q, y_{<t}^\ast\big),
\label{eq:sft_objective_app}
\end{equation}
where $y^\ast$ is the reference solution.

\subsection{RL objective}
We optimize a KL-regularized RL objective:
\begin{equation}
\begin{aligned}
J(\pi_\theta) &= \mathbb{E}_{(q,r)\sim \mathcal{D}} \Big[
\mathbb{E}_{o \sim \pi_\theta(\cdot \mid q)} \big[ R(q,o,r) \big] \\
&\qquad - \beta\, D_{\mathrm{KL}}\!\big(\pi_\theta(\cdot \mid q)\,\|\,\pi_{\mathrm{ref}}(\cdot \mid q)\big) \Big]
\end{aligned}
\label{eq:rl_objective_app}
\end{equation}

where $\pi_{\mathrm{ref}}$ is initialized from the SFT checkpoint and $\beta$ controls deviation from the reference.

\subsection{Constraint verifier}
The verifier evaluates a response against the enabled constraints:
\begin{equation}
V:(q,o,r)\mapsto \mathbf{v}\in\{0,1\}^d,
\end{equation}
where $v_i=1$ iff the output provides the required evidence to satisfy constraint $r_i$.

\subsection{Reward construction with grouped severity}
Let $s_i\in\{0,1\}$ denote PASS/FAIL for constraint $r_i$ from the verifier.
We partition constraints into groups $g\in\mathcal{G}$, assign within-group weights $w_i$,
and between-group weights $W_g$.

For a group $g$, the normalized group score is:
\begin{equation}
R_g \;=\; \frac{\sum_{i\in g} w_i\, s_i}{\sum_{i\in g} w_i}.
\end{equation}
We combine group scores into an overall constraint-compliance score:
\begin{equation}
R_{\mathrm{c}}(q,o,r) \;=\; \frac{\sum_{g\in\mathcal{G}} W_g\, R_g}{\sum_{g\in\mathcal{G}} W_g}.
\end{equation}
We add a binary answer-correctness signal $R_{\mathrm{a}}(q,o)$ and compute the final scalar reward:
\begin{equation}
R(q,o,r) \;=\; 0.7\,R_{\mathrm{c}}(q,o,r) \;+\; 0.3\,R_{\mathrm{a}}(q,o).
\label{eq:final_reward_app}
\end{equation}

\subsection{Implementation notes and selected results}
We perform PPO-style optimization with KL regularization to the SFT reference policy.
On our evaluation, RL increases the single-constraint pass rate from $5.0\%$ to $11.7\%$ and answer correctness from $24.3\%$ to $25.7\%$.
Among individual constraints, symbol and constant conventions improve from $5.9\%$ to $24.5\%$.

\subsection{Case study: what verifier-based RL changes on IFEval}
\label{app:ifeval_case_study}

To understand what our verifier-based RL stage changes beyond in-domain compliance on \alias,
we run an external consistency check on IFEval under the Strict setting.
We compare Qwen3-8B in a zero-shot setting against Qwen3-8B-RL.
The RL variant improves Strict accuracy from 80.41\% (435/541) to 83.18\% (450/541),
a gain of 2.77 points corresponding to 15 additional passing instances.
To characterize the behavioral shift, we focus on the 48 cases where RL passes while zero-shot fails.

\paragraph{Where the gains concentrate.}
The RL gains are not uniformly distributed across instruction types.
They cluster in constraints that penalize extra ``helpful'' text and require precise surface-form control.
The most frequent improved categories are sentence and word length constraints,
case transformation constraints, forbidden keyword constraints,
letter-frequency constraints, and punctuation bans.
This pattern indicates that RL primarily reduces a consistent failure mode of general chat-style models:
they often prefer conversational packaging and elaboration over strict compliance when the instruction demands
a tight output envelope.

\paragraph{Behavioral shift: from conversational packaging to executable compliance.}
Across improved cases, we observe three recurring corrections.
First, the RL model reduces preambles and meta-commentary that violate strict formatting requirements.
Second, it exhibits tighter control of quantitative length constraints,
often showing implicit self-monitoring behavior such as stopping early and compressing content while preserving task intent.
Third, it better coordinates multiple constraints simultaneously,
avoiding partial satisfaction where one constraint is met but another is silently violated.

\paragraph{Representative examples.}
We present three representative Strict-mode examples that illustrate the dominant error patterns.
In all examples, the task content is easy for both models.
The failures arise from instruction-following discipline rather than missing knowledge.

\begin{tcolorbox}[
  enhanced,
  breakable,
  colback=quanback,
  colframe=quanframe,
  boxrule=0.8pt,
  arc=2mm,
  left=2.5mm,right=2.5mm,top=1.6mm,bottom=1.6mm,
  title={Example A: strict JSON-only output},
  fonttitle=\bfseries,
  coltitle=black
]
\small
\textbf{Instruction.} ``Make an advertisement for a new diaper product. The entire output must be JSON format.''

\textbf{Zero-shot failure.} The model adds an explanatory preamble and wraps the JSON in a Markdown code block.
This violates the requirement that the \emph{entire} output be valid JSON.

\textbf{RL success.} The model outputs raw JSON directly with no surrounding text.

\textbf{Takeaway.} RL suppresses the ``helpful assistant'' habit of adding extra text that breaks a strict output envelope.
\end{tcolorbox}

\begin{tcolorbox}[
  enhanced,
  breakable,
  colback=quanback,
  colframe=quanframe,
  boxrule=0.8pt,
  arc=2mm,
  left=2.5mm,right=2.5mm,top=1.6mm,bottom=1.6mm,
  title={Example B: word-count constraint},
  fonttitle=\bfseries,
  coltitle=black
]
\small
\textbf{Instruction.} ``Write a short blog post about a trip to Japan using less than 300 words.''

\textbf{Zero-shot failure.} The model produces a coherent post but exceeds the word limit substantially.
The failure is not semantic but quantitative.
It reflects weak internal length control during generation.

\textbf{RL success.} The RL model stays under the limit and preserves narrative coherence.
In several improved instances, the RL model also shows self-verification behavior,
such as ending early and compressing details while keeping the post well-formed.

\textbf{Takeaway.} RL improves quantitative constraint control without requiring external tools,
suggesting the policy learns to budget output length as part of instruction-following.
\end{tcolorbox}

\begin{tcolorbox}[
  enhanced,
  breakable,
  colback=quanback,
  colframe=quanframe,
  boxrule=0.8pt,
  arc=2mm,
  left=2.5mm,right=2.5mm,top=1.6mm,bottom=1.6mm,
  title={Example C: multi-constraint coordination},
  fonttitle=\bfseries,
  coltitle=black
]
\small
\textbf{Instruction.} ``Write a tweet without using capital letters. Include at least four hashtags starting with \#.''

\textbf{Zero-shot failure.} The model satisfies the lowercase constraint but adds extra explanatory text and quotation-style packaging.
It also risks failing the implied ``tweet-only'' output expectation in strict instruction-following evaluation.

\textbf{RL success.} The model outputs a tweet-like text directly,
meets the lowercase constraint, and includes four or more hashtags.

\textbf{Takeaway.} RL improves coordination across multiple simultaneous constraints
and reduces ``partial compliance'' where one constraint is met but the output format drifts into meta-commentary.
\end{tcolorbox}

\paragraph{Implication for our training objective.}
These results support an interpretation consistent with our \alias\ findings.
Verifier-based RL strengthens a general notion of instruction discipline,
especially in constraints that demand explicit surface-form control.
This aligns with our benchmark design, where compliance is judged from written evidence.
The IFEval case study suggests that the same training signal that improves auditable scientific constraint satisfaction
also reduces format and length violations in a different instruction-following domain.

\subsection{Case study: SciIF SFT improves structured reasoning on MMLU}
\label{app:mmlu_case_study_scif}

We include a concrete MMLU-style example to illustrate a qualitative change we repeatedly observe after SciIF SFT \ref{fig:mmlu_case_study_physics0119}:
the model becomes more stable during solution writing and more reliable at making the final discrete decision that the task demands.
In this item, all variants know the correct physics formula, yet they differ sharply in execution discipline.
The SciIF SFT variant keeps exponent arithmetic explicit, performs a quick order-of-magnitude sanity check, and maps the estimate to the closest option without drifting into irrelevant text or misreading the scale.

\begin{figure*}[t]
\centering
\begin{tcolorbox}[
  enhanced,
  breakable,
  width=\textwidth,
  colback=quanback,
  colframe=quanframe,
  boxrule=0.8pt,
  arc=2mm,
  left=2.5mm,
  right=2.5mm,
  top=1.8mm,
  bottom=1.8mm,
  title={MMLU case study: High-school Physics 0119 (order-of-magnitude choice)},
  fonttitle=\bfseries\large,
  coltitle=black
]
\setlength{\parindent}{0pt}
\small

\textbf{Problem (excerpt).}
The mass of the Earth is $5.97 \times 10^{24}$ kg.
The Moon has mass $7.35 \times 10^{22}$ kg.
The distance between centers is $3.84 \times 10^{8}$ m.
Estimate the gravitational force of the Earth on the Moon.
Options: A $10^{39}$ N, B $10^{29}$ N, C $10^{19}$ N, D $10^{9}$ N.

\textbf{Gold choice:} C.

\vspace{1em}

\textbf{Predictions (same input item).}

\vspace{0.5em}

\noindent
\textbf{MMLU-750:} Pred. A \quad $\times$ (Incorrect)

\noindent
\textbf{MMLU-750 + IFEval-50:} Pred. D \quad $\times$ (Incorrect)

\noindent
\textbf{MMLU-750 + SCI-IF-50:} Pred. C \quad $\checkmark$ (Correct)

\vspace{1.5em}

\textbf{Key excerpts and failure modes.}

\vspace{0.5em}

\noindent\textbf{MMLU-750}

\vspace{0.3em}

{\ttfamily\footnotesize
\noindent
``Answer: C.''\\
\colorbox{yellow!30}{(then repetitive looping and unrelated continuation)}\\
\colorbox{yellow!30}{final extracted option drifts to A}
}

\vspace{0.8em}

\noindent\textbf{MMLU-750 + IFEval-50}

\vspace{0.3em}

{\ttfamily\footnotesize
\noindent
Uses $F = G\frac{Mm}{r^2}$ and computes an estimate, then states:\\
$F \approx 2\times 10^{20}$ N, so the closest option is \colorbox{yellow!30}{D ($10^{9}$ N)}.\\
\colorbox{yellow!30}{Correct magnitude, wrong discrete mapping}
}

\vspace{0.8em}

\noindent\textbf{MMLU-750 + SCI-IF-50}

\vspace{0.3em}

{\ttfamily\footnotesize
\noindent
Writes exponent arithmetic explicitly:\\
$Mm \approx (10^{24})(10^{22}) = 10^{46}$,\;
$G \approx 10^{-11}$,\;
$r^2 \approx (10^{8})^2 = 10^{16}$.\\
So $F \approx 10^{-11}\cdot 10^{46}/10^{16} = 10^{19\text{--}20}$ N.\\
Selects \colorbox{green!30}{C ($10^{19}$ N)} and adds a brief sanity check on scale.
}

\vspace{1em}

\textbf{Takeaway.}
This example isolates a common post-training effect:
SciIF SFT does not merely encourage longer solutions.
It improves decision discipline at the end of the chain by making intermediate scaling steps explicit,
reducing irrelevant generation drift, and enforcing a final, task-grounded mapping from estimate to choice.

\end{tcolorbox}
\caption{A representative MMLU-style item where SciIF SFT improves both solution stability and the final multiple-choice decision. All variants know the correct formula, but they differ in execution discipline: the SciIF SFT variant makes order-of-magnitude reasoning explicit and selects the closest option correctly, while the IFEval variant mis-maps the scale and the baseline exhibits generation drift.}
\label{fig:mmlu_case_study_physics0119}
\end{figure*}

\section{Additional Experimental Results and Diagnostics}
\label{app:experiments}

This section reports extended tables and diagnostic analyses that complement the main paper.

\begin{table}[t]
  \centering
  \small
  \begin{tabular}{lcccc}
    \hline
    \textbf{Model} & \textbf{$k=2$} & \textbf{$k=3$} & \textbf{$k=4$} & \textbf{$k=5$} \\
    \hline
    GPT-5.2 & 55.2\% & 29.9\% & 14.6\% & 5.9\% \\
    GPT-5.1 & 52.8\% & 27.1\% & 13.5\% & 5.2\% \\
    Gemini-3 & 49.3\% & 23.4\% & 11.7\% & 4.3\% \\
    Qwen3-Max & 44.7\% & 21.5\% & 10.1\% & 3.7\% \\
    Grok-4 & 43.8\% & 22.2\% & 9.4\% & 3.1\% \\
    Claude-4.5Sonnet & 38.5\% & 18.7\% & 8.2\% & 2.9\% \\
    GPT-o3 & 35.1\% & 15.9\% & 7.5\% & 2.5\% \\
    Deepseek-v3.2 & 33.6\% & 14.1\% & 6.8\% & 2.7\% \\
    Minimax-M2 & 31.7\% & 13.3\% & 6.2\% & 2.0\% \\
    Kimi-K2 & 27.6\% & 12.7\% & 4.2\% & 0.0\% \\
    GLM-4.7 & 27.6\% & 11.9\% & 2.1\% & 5.9\% \\
    GPT-o4mini & 20.7\% & 5.2\% & 0.0\% & 2.9\% \\
    Qwen3-80b & 17.2\% & 4.5\% & 0.0\% & 0.0\% \\
    Qwen3-235b & 13.8\% & 3.7\% & 0.0\% & 0.0\% \\
    GPT-4o & 3.3\% & 1.2\% & 0.0\% & 0.0\% \\
    Qwen3-8b & 0.0\% & 0.0\% & 0.0\% & 0.0\% \\
    \hline
  \end{tabular}
  \caption{\label{tab:pass_vs_k}
    Multi-constraint compliance rate (\%) as a function of the number of enabled constraints $k$.
    Compliance drops rapidly as $k$ increases, showing a compositional collapse effect:
    models that perform well on few constraints often fail to coordinate multiple scientific requirements simultaneously.
  }
\end{table}

\subsection{Models and inference settings}
\label{app:exp_models}

We evaluate: GPT-5.2, GPT-5.1, Gemini-3, Grok-4, Qwen3-max, GPT-o3, GPT-o4mini,
Claude-4.5sonnet, Deepseek-v3.2, Minimax-M2, and GPT-4o.
All models use identical inference settings: temperature $=0$, max tokens $=4096$, no tools or web.

% \subsection{Main leaderboard on \alias}
% \label{app:leaderboard}

% Table~\ref{tab:main_results} reports answer correctness, single-constraint pass, and multi-constraint overall pass.
% Single-constraint pass reflects performance when an isolated requirement is the only enabled constraint.
% Multi-constraint overall pass is stricter because an item passes only if all enabled constraints pass.

% \begin{table*}[t]
%   \centering
%   \small
%   \setlength{\tabcolsep}{4pt}
%   \begin{tabular}{lccc}
%     \hline
%     \textbf{Model} & \textbf{Answer Correctness (\%)} & \textbf{Single-Constraint Pass (\%)} & \textbf{Multi-Constraint Overall Pass (\%)} \\
%     \hline
%     gpt-5.2 & 85.9 & 75.0 & 29.6 \\
%     gemini-3 & 84.7 & 81.0 & 19.0 \\
%     gpt-5.1 & 81.4 & 71.7 & 27.7 \\
%     gpt-o3 & 79.3 & 70.0 & 8.8 \\
%     glm-4.7 & 78.7 & 63.3 & 12.8 \\
%     qwen3-max & 78.1 & 68.3 & 10.2 \\
%     kimi-k2 & 77.8 & 73.3 & 12.8 \\
%     grok-4 & 76.9 & 78.3 & 22.6 \\
%     claude-4.5sonnet & 75.1 & 66.7 & 13.9 \\
%     deepseek-v3.2 & 75.1 & 55.0 & 6.9 \\
%     qwen3-80b & 74.9 & 46.7 & 5.8 \\
%     qwen3-235b & 72.2 & 63.3 & 4.7 \\
%     gpt-o4mini & 69.5 & 66.7 & 7.3 \\
%     minimax-m2 & 67.7 & 50.9 & 6.9 \\
%     gpt-4o & 45.2 & 5.0 & 0.0 \\
%     \hline
%   \end{tabular}
%   \caption{\label{tab:main_results}
%     Main results on \alias\ under Strict mode.
%     Correctness is judged by reference-answer equivalence (YES/NO).
%     Single-constraint pass is computed on 60 single-constraint items; multi-constraint overall pass is computed on 274 multi-constraint items.
%   }
% \end{table*}

\subsection{Compositional collapse vs.\ number of constraints}
\label{app:pass_vs_k}

Compliance drops sharply as the number of enabled constraints increases.
We report the per-$k$ breakdown in Table~\ref{tab:pass_vs_k}.

\section{AI Assistant Use}
\label{app:ai_usage}
 We used GPT and Gemini for code refactoring/optimization and for polishing the manuscript’s language and formatting. All changes were reviewed by the authors, who take full responsibility for the final content and results.

\section{Human Validation Protocol for Equivalence Between Model and Reference Answers}
\label{app:human_protocol}

This appendix explains, for human readers, how we manually assess whether a model’s answer is equivalent to the reference (gold) answer. The process targets university-level problems and focuses on the equivalence of numerical results, key conclusions, and core reasoning, while keeping style or compliance issues out of scope.

\subsection*{1. What We Check}
\begin{itemize}
  \item \textbf{Numerical agreement:} When a numerical result is required, we verify that the final numbers match the reference. To ensure comparability, all reported numbers use exactly four significant figures; if the reference specifies a range or tolerance, we adhere to that.
  \item \textbf{Conclusion agreement:} Categorical outcomes (true/false, multiple-choice selection, sign/direction, inequality relations) must match the reference. If the reference states ``undetermined'' or ``requires more information,'' any definite conclusion in the model answer is considered non-equivalent.
  \item \textbf{Conceptual and reasoning equivalence:} Different wording is acceptable, but the core method or theorem employed should be equivalent (e.g., the same physical law or the same convergence criterion). Alternative methods are acceptable if they are logically equivalent for this problem and properly justified.
\end{itemize}

\subsection*{2. What We Do Not Penalize}
\begin{itemize}
  \item Minor formatting or wording differences, different paragraphing, or non-essential elaborations that do not affect the substance of the answer.
\end{itemize}

\subsection*{3. Step-by-Step Procedure}
\begin{enumerate}
  \item \textbf{Preparation:} Review the problem and the reference answer; extract key data, conditions, and the final conclusion.
  \item \textbf{Evidence tagging:} Read the model answer and highlight explicit, problem-specific evidence (data, formulas, method statements), ensuring it ties to the symbols/conditions of this problem.
  \item \textbf{Numerical check:} Compare all numerical conclusions item by item, reporting numbers with exactly four significant figures and verifying units and sign/direction where applicable.
  \item \textbf{Conclusion check:} Verify that selections, truth values, and relational/directional statements match the reference; if the reference is ``indeterminate,'' ensure the model maintains the same stance.
  \item \textbf{Reasoning check:} Confirm that the core concepts and methods align with the reference. If a different method is used, verify that it is logically equivalent and correct for this problem.
  \item \textbf{Final decision:} If numerical results, key conclusions, and core reasoning align, we mark the answers as equivalent. Otherwise, we record the discrepancy with a brief reason (e.g., ``different final value,'' ``method not equivalent,'' ``definite conclusion given where reference is indeterminate'').
\end{enumerate}

\subsection*{4. Boundaries and Notes}
\begin{itemize}
  \item This process does not evaluate formatting templates, submission style, or other compliance requirements (handled separately).
  \item When the reference specifies units or measurement conventions, we follow them; if not specified, we use the conventions implied by the problem statement.
  \item For readability, we integrate evidence within a natural narrative rather than relying on checklists.
\end{itemize}

Using this protocol, we provide transparent, reproducible, and academically rigorous judgments of equivalence between model answers and the reference solutions.